\documentclass[runningheads]{llncs}
\usepackage{graphicx}
\usepackage{amsmath}
\usepackage{amssymb}
\usepackage{cleveref}
\usepackage{tabularx}
\usepackage{url}
\begin{document}
\title{Kernel Corrector LSTM \thanks{This work was partially funded by projects AISym4Med (101095387) supported by Horizon Europe Cluster 1: Health,  ConnectedHealth (n.º 46858), supported by Competitiveness and Internationalisation Operational Programme (POCI) and Lisbon Regional Operational Programme (LISBOA 2020), under the PORTUGAL 2020 Partnership Agreement, through the European Regional Development Fund (ERDF); NextGenAI - Center for Responsible AI (2022-C05i0102-02), supported by IAPMEI,  and also by FCT plurianual funding for 2020-2023 of LIACC (UIDB/00027/2020\_UIDP/00027/2020) and SONAE IM Labs@FEUP.}}
%
%
\author{Rodrigo Tuna\inst{1}\orcidID{0009-0009-0047-2052} \and
Yassine Baghoussi\inst{1,4}\orcidID{0000-0002-1943-1471} \and
Carlos Soares\inst{1,2,3}\orcidID{0000-0003-4549-8917}\and
João Mendes-Moreira\inst{1,4}\orcidID{0000-0002-2471-2833}}
\authorrunning{R. Tuna et al.}
%
\institute{Faculdade de Engenharia, Universidade do Porto \\ \email{up201904967@edu.fe.up.pt}, \email{\{baghoussi,csoares,jmoreira\}@fe.up.pt}
\and
Artificial Intelligence \& Computer Science Lab.(LIACC -- member of LASI LA), Universidade do Porto
\and
Fraunhofer AICOS Portugal
\and
INESC TEC, Portugal}
\maketitle              
\begin{abstract}
Forecasting methods are affected by data quality issues in two ways: 1. they are hard to predict, and 2. they may affect the model negatively when it is updated with new data. The latter issue is usually addressed by pre-processing the data to remove those issues. An alternative approach has recently been proposed, Corrector LSTM (cLSTM), which is a Read \& Write Machine Learning (RW-ML) algorithm that changes the data while learning to improve its predictions. Despite promising results being reported, cLSTM is computationally expensive, as it uses a meta-learner to monitor the hidden states of the LSTM. We propose a new RW-ML algorithm, Kernel Corrector LSTM (KcLSTM), that replaces the meta-learner of cLSTM with a simpler method: Kernel Smoothing. We empirically evaluate the forecasting accuracy and the training time of the new algorithm and compare it with cLSTM and LSTM. Results indicate that it is able to decrease the training time while maintaining a competitive forecasting accuracy.

\keywords{Time series forecasting  \and Recurrent Neural Networks \and Data-Centric AI}
\end{abstract}
\section{Introduction}

In many fields, including energy, healthcare, management, and climate research, time series forecasting is a crucial task that can be accomplished using machine learning or statistical methods~\cite{introd}. As data becomes widely available, more precise forecasting models are expected. However, data quality issues like outliers, missing values, and changes in the underlying data generation process might impact predictive techniques.

Traditional machine learning (ML) models are often considered read-only models, capable of learning from data but neglecting the feedback loop for correcting the data during the learning process. This approach, while efficient in many cases, lacks proper adaptation of preprocessing techniques and the ML model itself, as the model's feedback is often overlooked. 

To address this limitation, the concept of Read-Write Machine Learning (RW-ML) has emerged. RW-ML models, such as Corrector LSTM (cLSTM)~\cite{clstm}, not only learn from data but also have the capability to change the data during the learning process. cLSTM is a time series forecasting method designed to improve forecasting accuracy by dynamically adjusting the data. It utilizes a meta-model of the Hidden State Dynamics obtained with SARIMA to detect data quality issues and employs a greedy heuristic to correct them. cLSTM has demonstrated superior predictive performance compared to traditional LSTM models. However, the computational cost associated with the meta-learning component of cLSTM is significant.

In this paper, we propose a computationally less expensive variant of cLSTM, named Kernel Corrector LSTM (KcLSTM), which replaces the meta-learner with a simpler method: Kernel Smoothing. We empirically compare KcLSTM with both cLSTM and LSTM models. Results reveal that KcLSTM achieves better predictive performance than LSTM and cLSTM, while also being faster than cLSTM, although the computational efficiency improvement is not as substantial as expected.

The main contributions of this paper are:
\begin{itemize}
    \item Introducing a variant of cLSTM, KcLSTM, which is computationally less expensive while maintaining high predictive accuracy.
    \item An empirical study comparing KcLSTM with LSTM and cLSTM in terms of predictive performance and training time.
\end{itemize}

This paper is structured as follows: we first provide an overview of the state-of-the-art forecasting method, LSTM. Then, we delve into the concept of RW-ML and its significance in time series forecasting. Next, we introduce the proposed algorithm, KcLSTM. Finally, we describe the experimental setup, present the results, and discuss their implications.

\section{Related Work}

In this section, we first present the Long Short-Term Memory. The algorithm that our proposed algorithm is built on and the one it will be compared to. We then define Data-Centric AI and provide examples of Data-Centric models built for time series forecasting.

\subsection{LSTM}
\begin{figure}
\centering
\includegraphics[width=0.7\textwidth]{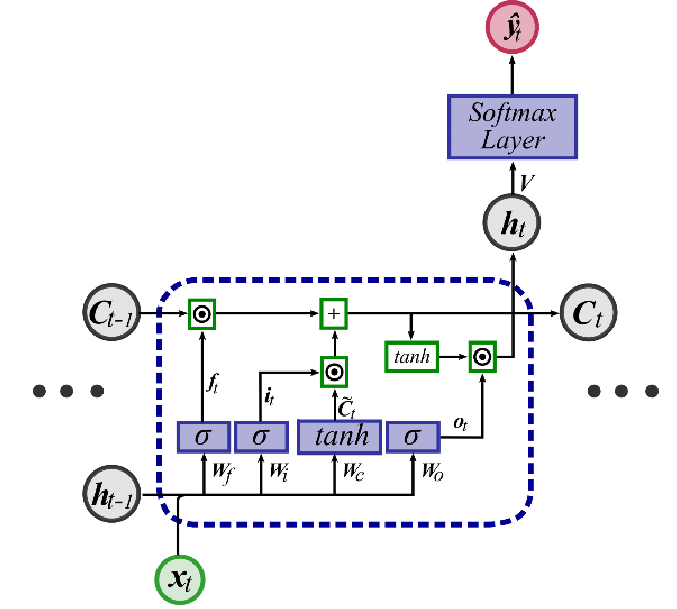}
\caption{Cell unit of the LSTM recurent neural network~\cite{lstmpic}.} \label{fig:lstm}
\end{figure}

The Long Short-Term Memory (LSTM) \cite{lstm} is a Recurrent Neural Network (RNN), that can capture long-term dependencies in the input and it is used for processing sequential data. RNNs differ from feed-forward networks through recurrent connections, allowing them to learn from sequential data. Back Propagation is applied to RNNs by taking advantage of the fact that for every recurrent network, there exists an equivalent feed-forward network with identical behavior for a finite number of steps \cite{rnn-lstm}, training it using Back Propagation Through Time (BPTT) \cite{lstm-tutorial}.

RNNs have some well-known limitations. First, they have problems capturing long-term dependencies, being limited to only bridge between 5-10 steps~\cite{lstm-tutorial}. This occurs because RNNs are sensible to the exploding/vanishing gradient problem~\cite{grad}. The LSTM solves this problem through the use of a gating mechanism.

An LSTM network consists of blocks, with each block containing an input gate, forget gate, output gate, and memory cell~(\cref{eq2,eq3,eq4}). The input gate controls which inputs are relevant; the forget gate learns which information should be kept in memory; and the output gate controls which information should be passed to the next block. The information is retained through the use of two states called the cell state, ~(\cref{eq5}), and the hidden state, ~(\cref{eq6}). The forward pass concatenates the input with the hidden state from the last block, while the backward pass derives the error and updates the gates using the chain rule of derivatives. The gate derivatives are multiplied by the hidden output to obtain the gradient deltas that update the gates.

\begin{subequations}
\begin{align}
\label{eq1}
    i_t & = \sigma(W_i \cdot h_{t-1} + V_i \cdot x_t + b_i) \\
\label{eq2}
    o_t & = \sigma(W_o \cdot h_{t-1} + V_o \cdot x_t + b_o) \\
\label{eq3}
    f_t & = \sigma(W_f \cdot h_{t-1} + V_f \cdot x_t + b_f) \\
\label{eq4}
    \hat{C_t} & = tanh(W_c \cdot h_{t-1} + V_c \cdot x_t + b_c) \\
\label{eq5}
    C_t & = i_t \cdot \hat{C_t} + f_t \cdot C_{t-1} \\
\label{eq6}
    h_t & = o_t \cdot tanh(C_t) \\
\label{eq7}
    z_t & = h_t
\end{align}
\end{subequations}

\subsection{Data-Centric Time Series Forecasting}

Anomalies, including outliers, missing values, and changes in the underlying data generation process can impact predictive tasks. This affects the predictions of such methods, hindering their performance~\cite{anomalies}, conversely to traditional machine learning methods, that build models using a fixed dataset. In Data-Centric AI~\cite{datacentric}, the focus is on the data. Data quality is increased to improve the performance of AI models.

The exploration into machine learning models capable of learning and correcting data has been a topic of interest in various studies. Both \cite{Song2017MachineLM} and \cite{Otte2013SafeAI} delve into this concept, with \cite{Song2017MachineLM} focusing on the potential of ML models to memorize sensitive information while \cite{Otte2013SafeAI} emphasize the importance of model interpretability and safety. Additionally, authors in \cite{Bhowmik2021ADA} further underscore the significance of data quality in enhancing model performance, advocating for a data-centric approach. Providing a broader perspective, the work in \cite{Ghahramani2015ProbabilisticML} discusses the role of probabilistic modeling in understanding learning and uncertainty in machine learning.

Moving to neural network models, authors in \cite{Denker1986NeuralNM} discuss highly interconnected networks for associative memory and optimization, with a focus on learning and adaptation. Moreover, \cite{Dehaene1987NeuralNT} propose a model for neural networks that learn temporal sequences through selection, employing synaptic triads and a local Hebbian learning rule. Furthermore, \cite{Dave2017PredictiveCorrectiveNF} introduce predictive-corrective networks for action detection in videos, which utilize top-down predictions and bottom-up observations for adaptive computation and simplified learning. These models collectively demonstrate the potential of neural networks to learn and correct data across various applications.

Similarly, recurrent neural network (RNN) models have been developed to address the challenge of learning and correcting data. In \cite{Gelenbe1992LearningIT}, an attempt is made to introduce a learning algorithm for the recurrent random network model, employing gradient descent of a quadratic error function. Later, authors in \cite{BailerJones1998ARN} propose a recurrent network architecture for modeling dynamical systems, which can learn from multiple temporal patterns and cope with sparse data. More recently, the research in \cite{Bowman2014RecursiveNN} demonstrates that tree-structured recursive neural networks can learn logical semantics, including entailment and contradiction.

In the context of time series forecasting, some data-centric approaches have been employed. In dLSTM~\cite{dlstm}, the authors train the model on non-anomalous data and use the predictive errors to detect anomalies. The deviation from the normal state is measured through delayed prediction errors. The normal state can then be restored from several candidate values.
Following the idea of using the prediction errors to improve the quality of the data Pastprop was introduced~\cite{pastprop}. The responsibility for the training error is shared between the model parameters and the training data. The backpropagation of the derivatives is applied to the input, indicating the part of the input that caused the training error. 

\subsection{Corrector LSTM}
cLSTM~\cite{clstm} is an architecture that improves its predictive performance by reconstructing the data of the model. The architecture of the algorithm is based on the LSTM and a data correction component. This data correction component uses a meta-learner, SARIMA, to identify problems in the hidden states of an LSTM model. This is achieved by predicting the hidden states using SARIMA and if the difference between the predicted and the real hidden states is over a certain threshold they are considered anomalous. The anomalies detected in the hidden states are assumed to be caused by the data which is then reconstructed. The reconstruction of the data points is such that the difference between the predicted and the real hidden states falls under a certain threshold. The authors showed that analyzing the Hidden State Dynamics~\cite{hidden} of an LSTM can be used to detect anomalies in the training data and consequently improve the forecasting performance of the model. However, the data correction relies on a meta-learner which makes the algorithm computationally expensive.

\section{Kernel Corrector LSTM}
The architecture of the Kernel Corrector LSTM (KcLSTM) is the same as the cLSTM architecture, and the meta-learner used to detect problems in the learning is substituted by a simpler approach, kernel smoothing.

\subsection{Training}
The KcLSTM utilizes the hidden states learned during the training process to find and correct data points of the series.
The training of the KcLSTM is divided into three distinct phases. The first phase consists of training the data on a standard LSTM. This allows the hidden states to capture the information of the time series and to be indicative of problems in the data. We then perform the correction, which is comprised of a detection and a correction component. These two components find and correct errors in the data respectively, this phase is thoroughly explained in \cref{corr}. Finally, the LSTM is trained on the new data, learning a corrected time series, that can improve the predictions of the model.

\subsection{Data Correction}\label{corr}
The Data Correction phase of the algorithm is divided into two different components: the correction and the detection. These two components aim to find data points that worsen the learning of the model and change the data so that the learning process is improved and a better model is obtained. Each phase has a threshold $\delta_d$ and $\delta_c$. 
The hidden states of the last iteration of the first training phase, $H = {h_0,..., h_n}$ are used to find errors in the training data. cLSTM uses a meta-model that is computationally expensive to compute; our goal is to assess if a simpler method can obtain competitive results with less cost; the method selected for this purpose is Kernel Smoothing because states are estimated rather than predicted which makes it computationally. A new set of estimated hidden states $H' = {h'_0,..., h'_n}$ is calculated using Gaussian Kernel Smoothing of $H$ as described in \cref{ks}.   
\begin{equation}\label{ks}
    h'_i = \frac{\sum_{j \in [i-W/2, i+W/2], i\ne j} h_j * K(h_i, h_j)}{\sum_{j \in [i-W/2, i+W/2], i\ne j} K(h_i, h_j)}
\end{equation}
Where $K(h_i, h_j)$ is:
\begin{equation}
    K(h_i, h_j) = e^{\frac{\lVert h_i - h_j \rVert ^2}{2\sigma^2}}
\end{equation}

\begin{figure}
\includegraphics[width=\textwidth]{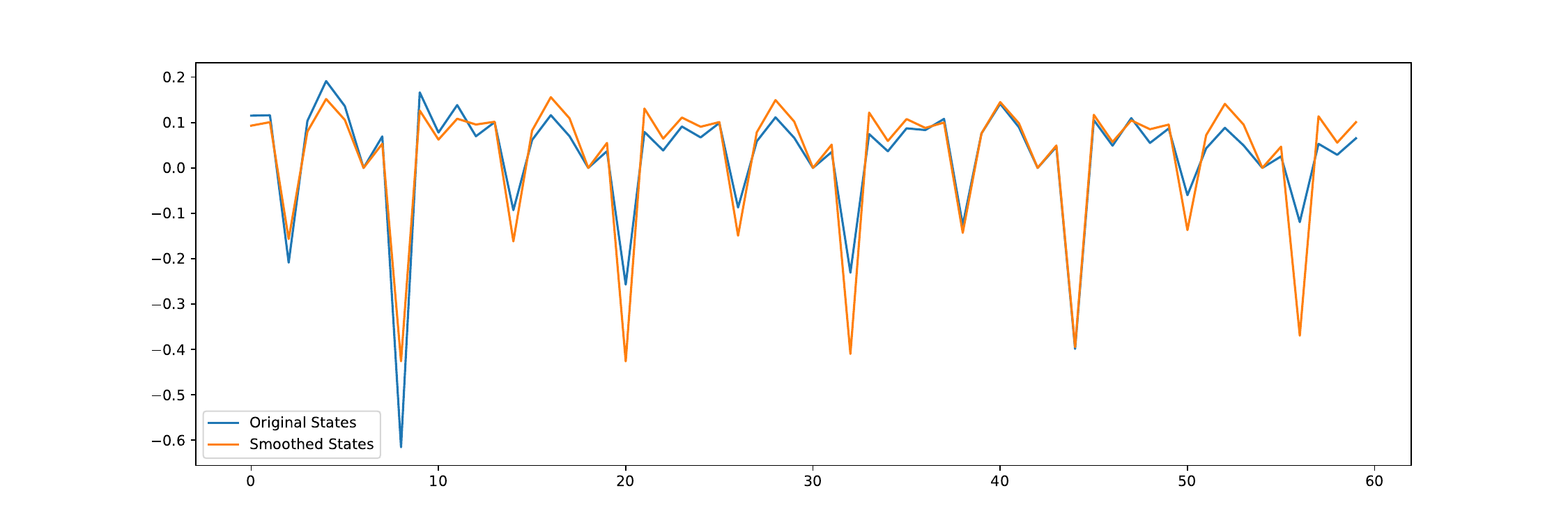}
\caption{Gaussian Smoothing of the Hidden States, represented as a series.} \label{fig1}
\end{figure}

The goal of error detection is to discriminate between data points that need reconstruction and those that do not. A point needs reconstruction if the Dynamic Time Warp similarity between the hidden state from which the point originated $h_i$ and the corresponding estimated hidden state $h'_i$ is greater than a given threshold $\delta_d$. This relation is depicted in \cref{detec}.
\begin{equation}\label{detec}
    DTW(h'_i, h_i) > \delta_d
\end{equation}

In the error correction, we reconstruct the detected points such that the Dynamic Time Warp similarity of the hidden state and estimated hidden state is less or equal to a given threshold $\delta_c$ \cref{rec}. Early stopping is employed, and if a maximum number of iterations is reached, the original value for the point is restored.
\begin{equation}\label{rec}
    DTW(h'_i, h_i) \le \delta_c
\end{equation}

\section{Experimental Setup}
The goals of the empirical validation are to investigate if the proposed algorithm is faster than the original one, without a significant decrease in forecasting accuracy.

A straightforward holdout method was used to estimate forecasting performance, used when there is a temporal dependency in the dataset~\cite{time-series}. The model is trained on the first $s$ samples and assessed on the succeeding $n-s$ samples. The data used for evaluation is always the original one. Using corrected data for the evaluation would likely lead to inadequate optimistic estimates of the forecasting performance of the corresponding method.

The hyperparameters of LSTM and KcLSTM are chosen using hyperparameter tuning using grid search. The learning rate was varied between: 0.0001, 0.001, 0.01, 0.1; and the batch size was varied between: 1, 2, 4, 8. For cLSTM we do not perform hyper-parameter tuning due to the high computational cost. Instead, we use the results described in the original paper~\cite{clstm}. The thresholds for KcLSTM are fixed with values of 0.6 for the detection and 0.5 for the correction. The thresholds for cLSTM are described in the original paper~\cite{clstm}, 0.6 for the detection and 0.2 for the correction. We chose to maintain the same detection threshold and increased the correction threshold. The kernel estimates of the hidden states are smoother; thus, a small correction threshold would significantly alter the hidden states, and the information learned in the previous phase would be lost. 

\subsection{Datasets}
\begin{table}
\caption{Statistical description of the dataset.}\label{tab1}
\begin{center}
\begin{tabular}{cc}
\hline    
   &  Monthly \\
  \hline
  Timeseries & 200 \\
  Average Length & 366 \\
  Mean &  4222 \\
  Standard Deviation & 1160\\
    \hline
\end{tabular}
\end{center}
\end{table}

We have used the M4 Competition Dataset~\cite{m4} comprising six subsets. From one subset, Monthly, we evaluate the performance of the algorithms on the first 199 time series. To evaluate the time taken to train the models, we use the first 20 time series of that subset.
\subsection{Evaluation Metrics}
This study focuses on both the predictive performance of the algorithm as well as its training time. To quantify the error of forecasts, we focus on the Mean Absolute Scaled Error (MASE) in Eq~\ref{mase} because it allows for the averaging of results across different time series as opposed to the Rooted Mean Squared Error (RMSE). The MASE measures the appropriateness of a forecast against the naive forecast of predicting the previous value. To assess if the differences are statistically different, we use the Mariano-Diebold Test~\cite{mariano}.
\begin{equation}\label{mase}
    MASE = \frac{\frac{1}{n-s}\sum_{i=s+1}^{n}{\hat{y}_i - y_i}}{\frac{1}{n-1}\sum_{i=2}^{n} |y_i - y_{i-1}|}
\end{equation}
The execution time is measured in seconds and the experiments were run on an Intel(R) Core(TM) i7-1065G7 CPU @ 1.30GHz processor.
\section{Results}
investigate if the proposed algorithm is faster than the original one, without significantly decreasing forecasting accuracy.
To illustrate the usefulness of the proposed algorithm, we first evaluate its forecasting capabilities, comparing it with LSTM~\cref{time} and cLSTM~\cref{predperf}. 

\subsection{Comparison with cLSTM}\label{predperf}
Results of MASE for the algorithms presented in~\cref{tab2} indicate KcLSTM outperforms cLSTM, but this may be explained by the hyper-parameter tuning that was performed for the KcLSTM and not the cLSTM. As such comparison between these two algorithms, can not be performed directly. 

The Mariano-Diebold Test for cLSTM and KcLSTM resulted in 40 wins for cLSTM, 111 wins for KcLSTM, and 48 draws. This shows an improvement in forecasting accuracy by substituting the meta-learner with Kernel Smoothing. Again the uneven conditions do not allow us to reach clear conclusions about these two methods.

Results for the training time presented in~\cref{tab2} indicate that KcLSTM is indeed faster than cLSTM, significantly. Nonetheless, the gain is not as large as would be expected. Estimating the states with Kernel Smoothing is less computationally expensive than predicting the states with SARIMA. However, this is a cruder method that results in estimated states that are farther away than from the original states when compared with. Consequently, more points are considered anomalies that will be corrected, which will cause the training time to increase.

\begin{table}
\caption{Comparison of each algorithm.}\label{tab2}
\begin{center}
\begin{tabular}{ c|c c c c }
\hline    
   & Mean & Median & Standard Deviation & Average time (s) \\
  \hline
  LSTM     &  3.48 & 0.74 & 6.07 & 20.47 \\
  cLSTM    & 8.77 & 1.04 & 36.96 & 56.15 \\
  KcLSTM   & 4.64 & 0.83 & 11.96 & 48.77  \\
    \hline
\end{tabular}
\end{center}
\end{table}

\begin{figure}
\includegraphics[width=\textwidth]{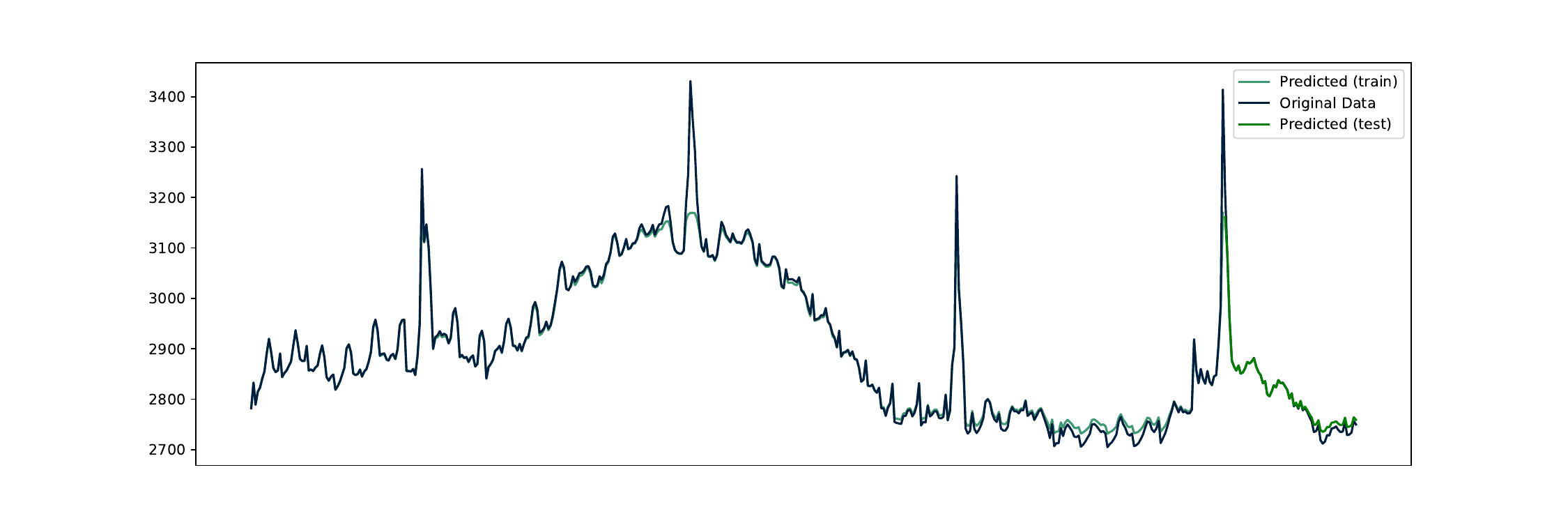}
\caption{Example where data reconstruction was successful.} \label{fig2}
\end{figure}

\begin{figure}
\includegraphics[width=\textwidth]{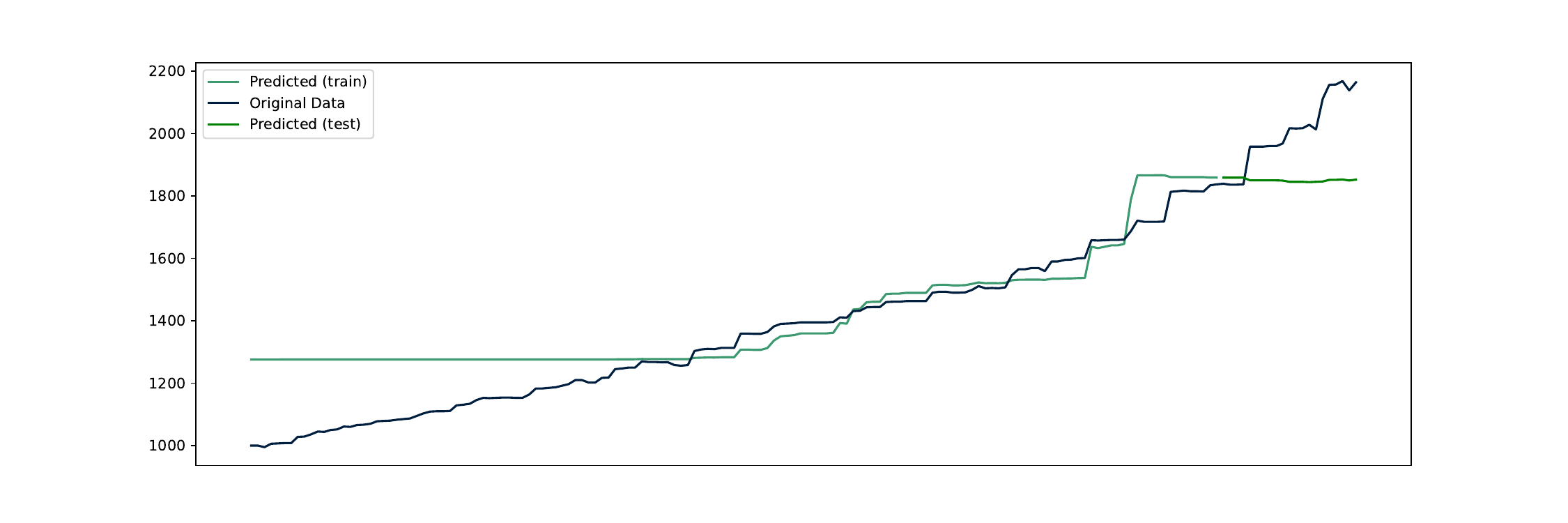}
\caption{Example where the data reconstruction destroyed the data which caused the model not to capture the information of the series.} \label{fig3}
\end{figure}

However, when performing the Mariano-Diebold test to compare LSTM and KcLSTM at the significance level of $\alpha = 0.05$ we get 102 wins for KcLSTM, 50 wins for LSTM, and 47 draws. We can conclude that KcLSTM is superior to LSTM as it wins more often, although when it loses it is by a greater margin. This is confirmed by the values of the standard deviation of MASE for LSTM and KcLSTM and explains the (apparent) superiority of LSTM when analyzing only the MASE. We see examples of a series with clear outliers that KcLSTM is able to correct and as such increase their predictions in~\cref{fig2}. Conversely, an example of a series without outliers made KcLSTM wrongfully alter the data which results in disastrous predictions in~\cref{fig3}. These two examples reflect the different behaviors mentioned before.

\subsection{Comparison with LSTM} \label{time}
Results of MASE for the algorithms presented in~\cref{tab2} indicate that LSTM has an overall better performance than KcLSTM with lower values for both the median and the mean for the MASE. However, when performing the Mariano-Diebold test to compare LSTM and KcLSTM at the significance level of $\alpha = 0.05$ we get 102 wins for KcLSTM, 50 wins for LSTM, and 47 draws. We can conclude that KcLSTM is superior to LSTM as it wins more often, although when it loses it is by a greater margin. This is confirmed by the values of the standard deviation of MASE for LSTM and KcLSTM and explains the (apparent) superiority of LSTM when analyzing only the MASE. We see examples of a series with clear outliers that KcLSTM is able to correct and as such increase their predictions in~\cref{fig2}. Conversely, an example of a series without outliers made KcLSTM wrongfully alter the data which results in worse predictions in~\cref{fig3}. These two examples reflect the different behaviors mentioned before.

Results for the training time presented in~\cref{tab2} indicate that KcLSTM is significantly slower than LSTM. This to be expected as KcLSTM has a data correction component that is responsible for most of the execution time of the algorithm.

\section{Conclusion}
The goal of our work is to create a forecasting algorithm that reconstructs data that is faster than current solutions in the literature.
We present a new algorithm: Kernel Corrector LSTM (KcLSTM). This algorithm alters the training data to improve its forecasting accuracy. Like in cLSTM, this is done by analyzing the Hidden States Dynamics and finding anomalies in hidden states to detect anomalies in data points and consequently correct them. However, the meta-learner of cLSTM was replaced by the Gaussian Kernel Smoothing of the hidden states to decrease the training time of cLSTM. 

We empirically compare our algorithm with LSTM and cLSTM both in terms of predictive performance and training time. Results show that KcLSTM obtains a competitive forecasting accuracy surpassing both the LSTM and cLSTM in the number of statistically significant wins. However, KcLSTM is more sensitive to its training data and more prone to making worse forecasts than the baseline, which caused the average MASE of LSTM to be inferior to the average MASE of KcLSTM. The measured training times also show that KcLSTM indeed improves on cLSTM in terms of computational cost, but the margin is smaller than expected because KcLSTM detects more points as anomalies than cLSTM. The estimated hidden states by KcLSTM are more distant from the real hidden states than the predicted states of cLSTM. The empirical study showed that KcLSTM is a faster algorithm that corrects its training data than cLSTM and that those corrections improve the forecasts by being superior to LSTM. Future work comprises the possibility of implementing the algorithm with different estimators.
\bibliographystyle{splncs04}
\bibliography{bibliography}
\end{document}